%% file: report.tex
\pdfoutput=1

\documentclass[11pt]{article}

\usepackage{ACL2023}

\usepackage{times}
\usepackage{latexsym}

\usepackage[T1]{fontenc}

\usepackage[utf8]{inputenc}

\usepackage{microtype}

\usepackage{inconsolata}

\usepackage{float}
\usepackage{amsmath}
\usepackage{amssymb}
\usepackage{hyperref}
\usepackage{graphicx}
\usepackage{algorithm}
\usepackage{algpseudocode}

\makeatletter
\@ifpackagewith{ACL2023}{review}{
    \newcommand{\noreview}[1]{}
}{
    \newcommand{\noreview}[1]{#1}
}
\makeatother

\title{Automatic Summarization of Long Documents}

\author{
  Naman Chhibbar \\
  IIT Hyderabad \\
  Kandi, Sangareddy \\
  Telangana 502285, India \\
  \texttt{ma21btech11011@iith.ac.in}
  \And
  Jugal Kalita \\
  University of Colorado, Colorado Springs \\
  1420 Austin Bluffs Pkwy \\
  Colorado Springs CO 80918 \\
  \texttt{jkalita@uccs.edu}
}

\begin{document}
  \maketitle

\input{sections/abstract}
\input{sections/introduction}
\input{sections/problem}
\input{sections/related-works}
\input{sections/datasets}
\input{sections/methodology}
\input{sections/metrics}
\input{sections/findings}
\input{sections/conclusion}

\input{anthology.bbl}
\end{document}

%% file: sections/abstract.tex
\begin{abstract}

A vast amount of textual data is added to the internet daily, making utilization and interpretation of such data difficult and cumbersome.
As a result, automatic text summarization is crucial for extracting relevant information, saving precious reading time.
Although many transformer-based models excel in summarization, they are constrained by their input size, preventing them from processing texts longer than their context size.
This study introduces three novel algorithms that allow any LLM to efficiently overcome its input size limitation, effectively utilizing its full potential without any architectural modifications.
We test our algorithms on texts with more than 70,000 words, and our experiments show a significant increase in BERTScore with competitive ROUGE scores.

\end{abstract}

%% file: sections/introduction.tex
\section{Introduction}
\label{sec:introduction}

Due to the ever-increasing amount of textual data available online, document summarization has become crucial for efficient and accurate extraction of relevant information.
Over the past few years, Large Language Models (LLMs) based on the transformer architecture \cite{vaswani2017attention} have shown ground-breaking abilities for NLP tasks, including document summarization \cite{yadav2023state}.
Recent developments have demonstrated remarkable improvements in the relevancy and coherence of summaries generated by such LLMs.

However, long document summarization, which makes reading, interpreting, and extracting information from vast texts accurate and efficient, remains a major challenge.
One of the major limitations in the transformer architecture is limited context size, stemming from the quadratic memory and computational complexity of the attention mechanism \cite{du2023improving}.
This limitation hampers the extraction of relevant information from lengthy texts, where summarization is essential to overcome the time, effort, and interpretive challenges posed by the complexity of such documents.

We experiment with three novel approaches to address the input size limitations of transformers.
The methods introduced do not include any architectural modifications and can be incorporated into any existing pipeline.
We believe that these methods can effectively utilize the full potential of any existing LLM.
Though our experiments only include the task of summarization, our methods can be applied to any NLP task (such as sequence classification, question-answering, and NLI) which requires processing long texts.

We start by stating the problem statement (\autoref{sec:problem}) and discussing related works (\autoref{sec:related-works}) to gain insights into the problem and the state-of-the-art solutions.
We then introduce the datasets (\autoref{sec:datasets}) and methodology (\autoref{sec:methodology}) used in our experiments.
For evaluating our results, we present some common metrics (\autoref{sec:metrics}) used in text summarization.
We end the report by discussing our experimental findings (\autoref{sec:findings}) and potential future work (\autoref{sec:future-work}), and concluding the study (\autoref{sec:conclusion}).

%% file: sections/problem.tex
\section{Problem Statement}
\label{sec:problem}

Our goal is to pre-process and manipulate a long document (with theoretically indefinite length) such that it fits within the context size of the model while retaining important information.
Practically, we have noticed that the document length may be up to \textbf{ten times} the context size of the model used.
For our experiments, we aim to reduce the summary length to about 400 words or less, preserving maximal salient information and coherence.

%% file: sections/related-works.tex
\section{Related Works}
\label{sec:related-works}

\citet{golia2024action} take a "Divide and Conquer" approach to address sequence length limitations in summarizing long meeting transcripts.
They begin by segmenting the transcript and then use the BART (Bidirectional and Auto-Regressive Transformer) \cite{lewis-etal-2020-bart} model to summarize each segment individually.
These segment summaries are then recursively combined and summarized until a single summary remains.
This method performs well with long documents but may take a considerable amount of time to converge due to repeated calls to the model.

There have also been efforts to improve the efficiency of the attention mechanisms in transformers.
\citet{beltagy2020longformer} introduce the Longformer, which replaces the quadratic self-attention mechanism in the Transformer architecture with a sliding window self-attention, resulting in a linear complexity with respect to the input size.
To capture long-range dependencies, they include global attention at specific token positions.
\citet{huang-etal-2021-efficient} modify the encoder-decoder attention mechanism such that each attention head in the decoder attends to $n/s_h$ tokens in the input sequence, where $n$ is the input length and $s_h$ is the number of heads.
This method has a complexity of $O(mn/s_h)$, where $m$ is the length of the output sequence.
\citet{bertsch2023unlimiformer} introduce Unlimiformer, which also modifies the encoder-decoder attention in a transformer.
The attention heads in the decoder only attend to the tokens picked by their k-Nearest-Neighbor (kNN) algorithm.
The kNN indices between the input tokens are created by the hidden states generated in the encoder.
\citet{phang2022investigating} introduce the staggered block-local attention mechanism.
In the block-local attention mechanism, the input sequence is divided into multiple non-overlapping blocks.
Tokens in a block attend only to the tokens in the same block.
In staggered block-local attention, the blocks are staggered such that each token is in a different block in each head.

Other unique approaches include VideoAgent, introduced by \citet{wang2024videoagent}, an AI agent designed to answer a given question based on a long video.
They achieve this by generating captions from multiple uniformly sampled frames from the video.
These captions are used to answer the user's question.
\citet{chen2022long} describe a novel algorithm to classify long Chinese news into a set of predefined categories.
They form multiple groups of sentences based on a maximum token threshold in each group.
These groups are then encoded using BERT (Bidirectional Encoder Representations from Transformers) \cite{devlin2018bert} and passed through a 1D convolution layer for local feature extraction.
What makes this method special is that the attention mechanism is replaced by a 1D convolution layer, which has linear complexity.
\citet{chen2023extending} use positional interpolation to extend the context size of a pre-trained model.
Instead of the usual extrapolation of the positional embeddings, they downscale the positional embeddings to force them into a range the model is trained on, hence interpolating in the pre-trained range.
They claim that the model should use the positional embeddings on which it is trained.

%% file: sections/datasets.tex
\section{Datasets}
\label{sec:datasets}

We use datasets containing documents with a maximum word count exceeding 70,000.
We briefly discuss and analyze the word counts of the datasets below.

\subsection*{GovReport}

\begin{figure}[!ht]
  \centering
  \includegraphics[width=.48\textwidth]{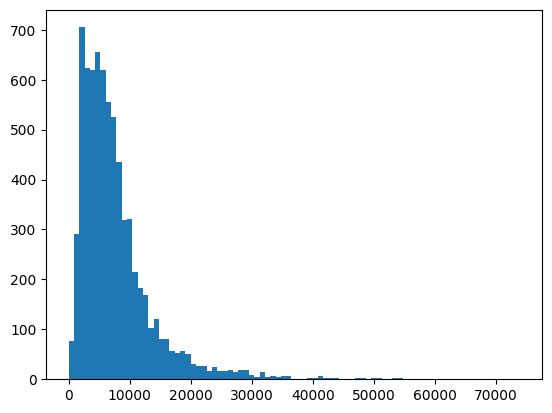}
  \caption{
    GovReport word counts.
    Document word counts are on the x-axis with the number of documents on the y-axis.
  }
  \label{fig:govreport}
\end{figure}

Introduced by \citet{huang-etal-2021-efficient}, this dataset consists of reports written by government research agencies, including the Congressional Research Service (CRS) and the U.S. Government Accountability Office (GAO).
Exact word count information is given in \autoref{tab:datasets}.
\autoref{fig:govreport} shows the word count distribution of the dataset.

\subsection*{BigPatent}

\begin{figure}[!ht]
  \centering
  \includegraphics[width=.48\textwidth]{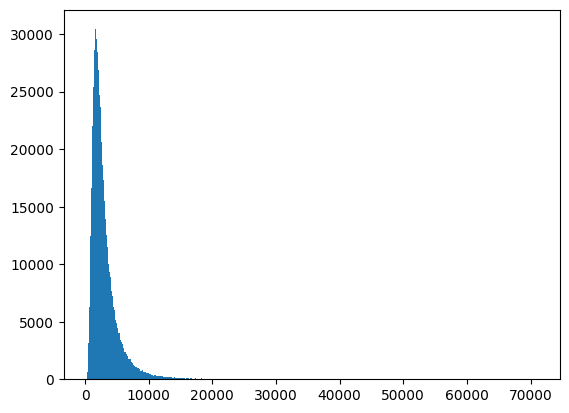}
  \caption{
    BigPatent word counts.
    Document word counts are on the x-axis with the number of documents on the y-axis.
  }
  \label{fig:bigpatent}
\end{figure}

Introduced by \citet{sharma-etal-2019-bigpatent}, this dataset consists of over 1.3 million records of U.S. patent documents with human-written abstractive summaries.
Exact word count information is given in \autoref{tab:datasets}.
\autoref{fig:bigpatent} shows the word count distribution of the dataset.

\begin{table*}[!ht]
  \centering

  \begin{tabular}{c c c c}
    \hline
    Dataset & Avg. Word Count & Max Word Count & No. of Documents \\
    \hline
    GovReport & 7,700.71 & \textbf{73,815} & 7,238 \\
    BigPatent & 3,055.72 & \textbf{71,027} & 1,341,362 \\
    \hline
  \end{tabular}

  \caption{Dataset information}
  \label{tab:datasets}
\end{table*}

%% file: sections/methodology.tex
\section{Methodology}
\label{sec:methodology}

In this section, we discuss the three algorithms used for distilling documents.
Two of our algorithms start by segmenting the document into smaller, contiguous, and exhaustive parts.
We do so by using a sentence tokenizer to separate sentences from the text and then merging them such that the number of words in each segment is more than the threshold $min\_words$, hyperparameter in both methods.

\subsection{Central Truncation}
\label{method:truncation}

Truncation is the most common and straightforward approach used to handle long texts that exceed the context size of an LLM.
It can be done in three main ways:

\begin{itemize}
  \item \textbf{Retaining Head}: Keeping tokens from the start.
  \item \textbf{Retaining Tail}: Keeping tokens from the end.
  \item \textbf{Head and Tail}: Keeping tokens from both start and end.
\end{itemize}

\citet{worsham-kalita-2018-genre} also employ "retaining head" and "retaining tail" strategies on long texts and find promising results for long text genre classification.
Though the "retaining head" method is often used, keeping the initial tokens allowed by the LLM, \citet{sun2019fine} find that keeping both head and tail produces better results than both the "retaining head" and the "retaining tail" methods.
Their research also shows that truncating the middle is even better than the more complicated hierarchical methods, displaying superiority with simplicity.
This is a time-efficient method worth exploring.

The fraction of tokens to be taken from the head is controlled by the hyperparameter $head\_size \in [0, 1]$ in our algorithm.
Setting $head\_size = 1$ results in taking tokens only from the head, whereas setting $head\_size = 0$ results in taking tokens only from the tail.
The truncated tokens are then sent to the model for summarization.

\subsection{Document Skimming}
\label{method:skimming}

\begin{figure*}[!ht]
  \centering
  \includegraphics[width=.8\textwidth]{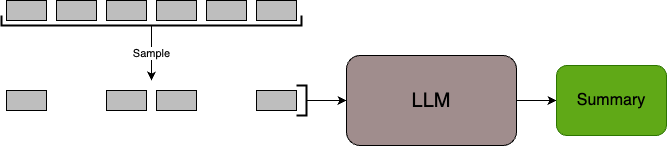}
  \caption{The Document Skimming Algorithm. The grey blocks represent segments of the document.}
  \label{fig:doc-skim}
\end{figure*}

One way to process long texts is by employing a speed reading strategy called skimming \cite{dhillon2020effect}.
Skimming is performed by reading the whole text in a go while selectively skipping some parts of the text for quicker reading.
The reader usually omits the portions that seem redundant or irrelevant in the text, minimizing information loss.
This method is inspired by the way \citet{wang2024videoagent} randomly sample video frames to generate captions.
\citet{worsham-kalita-2018-genre} also use random sampling for genre identification.

This method starts by segmenting the document with the hyperparameter $min\_words$ (introduced at the start of \autoref{sec:methodology}).
We then sample segments uniformly, with each segment having probability $p$ to be picked.
The sampled segments are then concatenated to form a single text and sent to the model.
This method ensures the model sees a segment from each part of the text.
\autoref{fig:doc-skim} is a visual representation of the algorithm.

Below is an example of the distilled text generated by the algorithm and the summary generated by GPT-3.5 Turbo \cite{brown2020language}:

\noindent \textbf{Example Text:}
\begin{quote}
  Title: Awards of Attorneys’ Fees by Federal Courts and Federal Agencies.
  Subsection I. Introduction: The American ...
\end{quote}

\noindent \textbf{Distilled Text:}
\begin{quote}
  Alyeska Pipeline Service Co. v. Wilderness Society , 421 U.S. 240, 247 (1975). This is known as the "American rule" (as opposed to the ...
\end{quote}

\noindent \textbf{Summary:}
\begin{quote}
  The American rule regarding attorneys' fees has two common law exceptions: the common benefit doctrine and bad faith ...
\end{quote}

\begin{figure}
  \centering
  \includegraphics*[width=.45\textwidth]{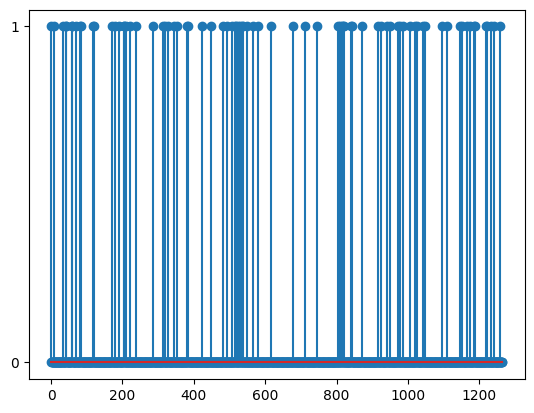}
  \caption{Segments picked by the Document Skimming algorithm. Y-axis value of the ith segment
  on x-axis is 1 if its picked, 0 otherwise.}
  \label{fig:uniform}
\end{figure}

Refer to \autoref{fig:uniform} to visualize the segments picked by the algorithm.

\subsubsection*{Removing Redundancy}

To address the issue of redundancy in the document, we experiment with and without removing redundant segments before and after sampling.
We do this to prevent the model from seeing the same information multiple times, which may lead to repetition in the output.
This is achieved by linearly iterating over the sampled segments and selectively removing some of the segments.
We do this by maintaining the mean embedding of the selected segments, initialized as a zero vector.
The current segment is retained if the cosine similarity between the mean embedding and the segment embedding is lower than a $threshold$, which acts as a hyperparameter.
A \href{https://huggingface.co/sentence-transformers/all-MiniLM-L6-v2}{sentence} transformer is used to generate the segment embeddings.
The sentence transformer is based on MiniLM \cite{wang2020minilm}, which is a distilled version of a larger encoder-only transformer model.
In case the current segment is retained, the mean embedding is updated as follows:

\[ new\_mean\_emb = \frac{n \cdot mean\_emb + seg\_emb}{n + 1} \]

\noindent where $n$ is the number of sampled segments (excluding the current segment),
$seg\_emb$ is the segment embedding of the current segment, $mean\_emb$ is the mean embedding, and $new\_mean$ is the updated mean embedding.

While removing segments after sampling, we waste some of the context size.
To alleviate this, we increase the probability of choosing a segment during sampling to compensate for the removed segments.
This fraction is controlled by the hyperparameter $prob\_boost$.
The updated probability is calculated as follows:

\[ p_{new} = (1 + prob\_boost) \cdot p \]

Even though removing redundant segments before sampling is less efficient due to the whole document being processed, it ensures better utilization of the LLM's context size.

\subsubsection*{Other Calculations}

We now discuss caluclation of the optimal value of $p$.
Let $X$ denote the total number of tokens in the sampled segments.
Since segments are sampled randomly, $X$ is a random variable.
If the context size of the model is $model\_size$, we want $\mathrm{E}[X] = model\_size$, where $\mathrm{E}[X]$ denotes the expectation of $X$.

Suppose we have $n \in \mathbb{N}$ segments and $X_i \sim \mathrm{Bernoulli}(p)$ denotes if segment $i$ is chosen, $i \in \{1, 2, \dots, n\}$.
If $len_i$ denotes the number of tokens in segment $i$, we can write:

\begin{align*}
  X &= \sum_{i = 1}^{n} X_i \cdot len_i \\
  \Rightarrow \mathrm{E}[X] &= \mathrm{E}[\sum_{i = 1}^{n} X_i \cdot len_i] \\
  &= \sum_{i = 1}^{n} \mathrm{E}[X_i \cdot len_i] \\
  &= \sum_{i = 1}^{n} \mathrm{E}[X_i] \cdot len_i
\end{align*}

Since $X_i \sim \mathrm{Bernoulli}(p)$ $\forall i \in \{1, 2, \dots, n\}$, we have $\mathrm{E}[X_i] = p$ $\forall i \in \{1, 2, \dots, n\}$.

\begin{align*}
  \therefore \mathrm{E}[X] &= \sum_{i = 1}^{n} p \cdot len_i \\
  &= p \cdot \sum_{i = 1}^{n} len_i
\end{align*}

Let $total\_len$ be the total number of tokens in the text, then $total\_len = \sum_{i = 1}^{n} len_i$.

\[ \therefore \mathrm{E}[X] = p \cdot total\_len = model\_size \]
\[ \Rightarrow p \cdot total\_len = model\_size \]
\[ \Rightarrow p = model\_size / total\_len \]

\subsection{Summarization with Keyword Extraction}
\label{method:keyword}

\begin{algorithm*}
  \caption{Summarization with Keyword Extraction}

  \begin{algorithmic}
    \State \textbf{Input:} $text$ (text), $size$ (context size of model)
    \State \textbf{Output:} Distilled text
    \State $segments \leftarrow \text{segmenter}(text)$
    \State $embeddings \leftarrow \text{sentence\_transformer}(segments)$
    \State $keywords \leftarrow \text{LDA}(text)$
    \State $\text{concatenate}(keywords, \text{delimiter})$
    \State $keyword\_embedding \leftarrow \text{sentence\_transformer}(keywords)$
    \State Sort $embeddings$ by decreasing cosine similarity scores with $keyword\_embedding$
    \State $selected \leftarrow \{\}$
    \State $num\_tokens \leftarrow 0$
    \For{$embedding \in embeddings$}
      \State $tokens \leftarrow \text{count\_tokens}(embedding)$
      \If{$tokens + num\_tokens \le size$}
        \State $selected \leftarrow selected \cup \{embedding\}$
        \State $num\_tokens += tokens$
      \EndIf
    \EndFor
    \State $\text{concatenate}(selected, \text{delimiter})$
    \State \Return{$selected$}
  \end{algorithmic}

  \label{algo:keyword}
\end{algorithm*}

Document skimming (\autoref{method:skimming}) involves a very intuitive and simple approach of sampling segments randomly.
In an attempt to use the entirety of the text, we experiment with an efficient keyword extraction algorithm to get important keywords that explain the core meaning of the document.
These keywords capture the overall meaning of the document and can help us sample segments intelligently, ensuring we get the most important segments from the document.

We use Latent Dirichlet Allocation (LDA) \cite{blei2003latent} with a single topic to get the topic words (or keywords) from the document.
There are many ways to use these to create a probability distribution to sample the segments.
A simple approach we use is to concatenate the keywords using a delimiter (a space is used in our experiments) to form a single sentence.
This sentence is then embedded to form the keyword embedding, which, in theory, captures a high-level meaning of the document.
The keyword sentence and document segments are embedded using the same \href{https://huggingface.co/sentence-transformers/all-MiniLM-L6-v2}{sentence transformer} used in the previous method.
The segment embeddings are then compared to the keyword embedding using cosine similarity to get similarity scores for each segment embedding.
The maximum possible number of segments with the highest similarity scores are retained.
The selected segments are then concatenated and sent to the model.
\autoref{algo:keyword} describes the process.

Below is an example of the distilled text generated by the algorithm and the summary generated by GPT-3.5 Turbo \cite{brown2020language}:

\noindent \textbf{Example Text:}
\begin{quote}
  Title: Awards of Attorneys’ Fees by Federal Courts and Federal Agencies. Subsection I. Introduction: The American ...
\end{quote}

\noindent \textbf{Distilled Text:}
\begin{quote}
  Title: Awards of Attorneys’ Fees by Federal Courts and Federal Agencies Subsection I. Introduction: The American Rule and ...
\end{quote}

\noindent \textbf{Summary:}
\begin{quote}
  The document discusses the American Rule regarding attorneys' fees, where prevailing litigants are not typically entitled to ...
\end{quote}

\begin{figure}
  \centering
  \includegraphics*[width=.45\textwidth]{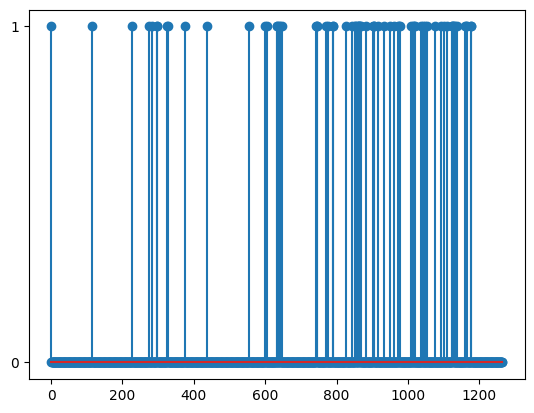}
  \caption{
    Segments picked by the Summarization with Keyword Extraction algorithm.
    Y-axis value of the ith segment on x-axis is 1 if its picked, 0 otherwise.
  }
  \label{fig:keyword}
\end{figure}

Refer to \autoref{fig:keyword} to visualize the segments picked by the algorithm.

This approach is similar to the way \citet{golia2024action} use action items to pick segments of text (a neighbourhood of 2 sentences around the action item) to obtain meeting minutes.

%% file: sections/metrics.tex
\section{Evaluation Metrics}
\label{sec:metrics}

The best way to evaluate generated natural language is by humans, but conducting human trials are expensive and time-consuming.
Hence, we use automatic evaluation metrics to evaluate the quality of the generated summary, given reference summaries.
\citet{fabbri2021summeval} review many such open-source and state-of-the-art metrics.
The two that we use in our experiments are discussed below.
These metrics are commonly used in published literature.

\textbf{ROUGE metrics:} \citet{lin-2004-rouge} introduces the Recall-Oriented Understudy for Gisting Evaluation (ROUGE) metrics.
The basic ROUGE-N metric is based on the fraction of overlaps of ideal or reference summaries with the candidate summary, hence being recall-oriented.
His study concludes that ROUGE-N with $\text{N} = 2$, ROUGE-L, ROUGE-W, and ROUGE-S work well for the summarization task.

\textbf{BERTScore:} \citet{zhang2019bertscore} introduce the BERTScore, an automatic evaluation metric for text generation.
BERTScore is calculated by comparing the contextual embeddings of tokens in the candidate and reference summaries, which are generated using BERT \cite{devlin2018bert}.
BERTScore excels at capturing semantic similarities between sentences since it uses contextual embeddings of tokens instead of using N-gram frequencies to calculate similarity.

%% file: sections/findings.tex
\section{Experimental Findings}
\label{sec:findings}

\begin{table*}[!ht]
  \centering

  \begin{tabular}{c c c c c}
    \hline
    Model & ROUGE-1 & ROUGE-2 & ROUGE-L & BERTScore \\
    \hline
    BART w/ Unlimiformer (1,024) & 53.4 & 22.5 & 22.5 & 66.0 \\
    PRIMERA w/ Unlimiformer (4,096) & 56.5 & 24.8 & 26.3 & 67.7 \\
    Hepos (10,240) & 51.34 & 19.09 & \textbf{48.73} & - \\
    PEGASUS-X w/ Staggered & 60.3 & \textbf{30.0} & 31.5 & - \\
    Block-Local Attention (16k) & & & & \\
    LLaMA-7B w/ Positional & 60.0 & 28.0 & 29.5 & - \\
    Interpolation (15k) & & & & \\
    \hline
    Summarization w/ Extraction & \textbf{61.99} & 18.52 & 38.46 & \textbf{86.20} \\
    + GPT-3.5 Turbo (4,096) & & & & \\
    Central truncation + LongT5 (4,096) & 46.20 & 4.38 & 38.27 & \textbf{82.19} \\
    Skimming w/ post-sampling & 46.76 & 4.56 & 39.61 & \textbf{81.96} \\
    removal + LongT5 (4,096) & & & & \\
    \hline
  \end{tabular}

  \caption{
    Automatic evaluation results on the GovReport dataset. Context size of the models are mentioned in parentheses.
    The best score in each metric category is highlighted in \textbf{bold}.
    Results of our algorithms are below the horizontal line in the middle.
  }
  \label{tab:govreport}
\end{table*}

\begin{table*}[!ht]
  \centering

  \begin{tabular}{c c c c c}
    \hline
    Model & ROUGE-1 & ROUGE-2 & ROUGE-L & BERTScore \\
    \hline
    BigBird-Pegasus (16k) & \textbf{60.64} & \textbf{42.46} & \textbf{50.01} & - \\
    \hline
    Skimming w/ pre-sampling & 27.40 & 3.31 & 21.25 & \textbf{82.62} \\
    removal + GPT-3.5 Turbo (4,096) & & & & \\
    Central truncation + GPT-3.5 Turbo (4,096) & 27.77 & 3.09 & 20.56 & \textbf{82.57} \\
    Skimming w/ post-sampling & 26.16 & 2.13 & 20.21 & \textbf{82.40} \\
    removal + GPT-3.5 Turbo (4,096) & & & & \\
    \hline
  \end{tabular}

  \caption{
    Automatic evaluation results on the BigPatent dataset. Context size of the models are mentioned in parentheses.
    The best score in each metric category is highlighted in \textbf{bold}.
    Results of our algorithms are below the horizontal line in the middle.
  }
  \label{tab:bigpatent}
\end{table*}

We test our pipelines with the following models: \textbf{BART} (Bidirectional and Autoregressive Transformer) \cite{lewis-etal-2020-bart} fine-tuned on the CNN/Daily Mail dataset \cite{nallapati2016abstractive} with a context size of 1024, \textbf{LongT5} \cite{guo2021longt5}, a variant of T5 (Text-to-Text Transfer Transformer) \cite{raffel2020exploring}, fine-tuned on the BookSum dataset with a context size of 4096, and \textbf{GPT-3.5 Turbo} \cite{brown2020language} with a context size of 4096.

We compare our results with the state-of-the-art summarization models on the GovReport dataset, including Unlimiformer \cite{bertsch2023unlimiformer} integrated with BART \cite{lewis-etal-2020-bart} and PRIMERA \cite{beltagy2020longformer}, Hepos \cite{huang-etal-2021-efficient}, PEGASUS-X with staggered block-local attention \cite{phang2022investigating}, extended	LLaMA-7B with positional interpolation \cite{chen2023extending}.
We also compare our results with BigBird-Pegasus \cite{zaheer2020big} on the BigPatent dataset.
Refer to \autoref{tab:govreport} and \autoref{tab:bigpatent} for results on the GovReport and BigPatent datasets, respectively.

We were unable to obtain the BERTScores of our baselines, except for Unlimiformer, due to unavailability of code or computational limitations.

\subsection*{Time complexity analysis}

\begin{figure}[!ht]
  \centering
  \includegraphics[width=.48\textwidth]{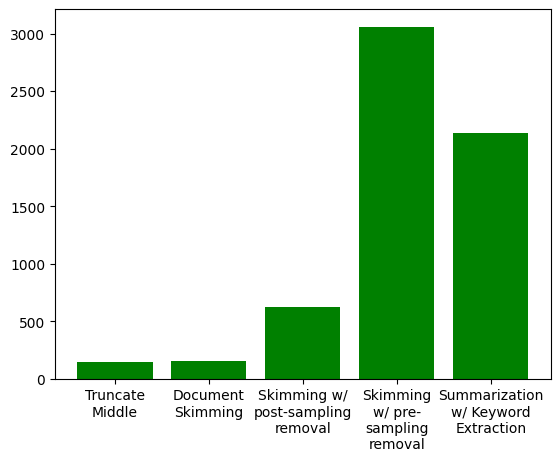}
  \caption{Mean time taken per document using BART tokenizer on BigPatent dataset}
  \label{fig:times}
\end{figure}

We evaluate the time complexity of our methods by measuring the mean time taken to process a document (excluding the time taken by the model to generate the summaries).
We find that Central Truncation (\autoref{method:truncation}) and Document Skimming (\autoref{method:skimming}) take approximately the same time.
Skimming with post-sampling removal takes slightly more time than the other two methods.
We can see a significant increase in time taken by Skimming with pre-sampling removal and Summarization with Keyword Extraction (\autoref{method:keyword}) due to the additional computations required.
\autoref{fig:times} illustrates the average time taken by our methods.
Check \autoref{tab:encoder-times} for exact values rounded off to two decimal places.

\begin{table}[!ht]
  \centering

  \begin{tabular}{c c}
    \hline
    Method & Mean time taken \\
    \hline
    Central Truncation & 142.50 ms \\
    Document Skimming & 155.42 ms \\
    Skimming w/ post- & 625.17 ms \\
    sampling removal & \\
    Skimming w/ pre- & 3059.63 ms \\
    sampling removal & \\
    Summarization & 2131.40 ms \\
    w/ Extraction & \\
    \hline
  \end{tabular}

  \caption{Mean time taken (in milliseconds) per document using BART tokenizer on BigPatent
  dataset}
  \label{tab:encoder-times}
\end{table}

%% file: sections/conclusion.tex
\section{Future Work}
\label{sec:future-work}

To segment the document, we use a basic sentence tokenizer (\href{https://www.nltk.org/api/nltk.tokenize.sent_tokenize.html}{nltk.sent\_tokenize}) with some modifications to control the minimum number of words in a segment.
In our experiments, we find that segmentation is a crucial step in the pipeline and can influence the output summary greatly, indicating that good segmentation is important for good distillation of text.
Ensuring the uniformity of the length of the segments while preserving coherence within a segment is also essential for better utilization of the context size of the model.
We encourage future work to experiment with different kinds of segmenters.

Future work may also be focused on extending the Summarization with Keyword Extraction (\autoref{method:keyword}) method.
There are many potential ways to use the extracted keywords we do not touch upon.

\section{Conclusion}
\label{sec:conclusion}

Our experiments show that Document Skimming with post-sampling removal (\autoref{method:skimming}) performs well while being efficient.
The Central Truncation method (\autoref{method:truncation}) also shows good results, which shows that simple methods can also be effective when dealing with long inputs.
The last two methods, Skimming with pre-sampling removal (\autoref{method:skimming}) and Summarization with Keyword Extraction (\autoref{method:keyword}), achieve the best results but are computationally expensive.

Our experiments show significant improvement in BERTScore compared to Unlimiformer \cite{bertsch2023unlimiformer} on the GovReport dataset.
This shows that our pipelines can utilize details in long texts efficiently.
Even though our ROUGE-2 scores are lower than the baselines, ROUGE-1 and ROUGE-L scores are competitive.
Since BERTScore is better at capturing semantic similarity, we highlight the use of BERTScore compared to ROUGE scores.
Hence, we hypothesize that our pipelines can generate better summaries than the baselines with higher ROUGE scores.
It should also be noted that the models used in our experiments have smaller context sizes compared to the baselines, indicating that our algorithms have a greater potential if used with larger models.

\noreview{
  \section*{Acknowledgement}
  
  All work herein reported is supported by the Nation Science Foundation under Grant No. 2349452.

  Any opinion, finding, or conclusion in this study is that of the authors and does not necessarily
  reflect the views of the National Science Foundation.
}

\section*{Supplementary Materials}

The datasets used in this study are available here:
\href{https://gov-report-data.github.io/}{GovReport},
\href{https://evasharma.github.io/bigpatent/}{BigPatent}

The code used in this study is available here:
\href{https://github.com/NamanChhibbar/Long-Document-Summarizer.git}{GitHub}